\documentclass{article}

\usepackage{PRIMEarxiv}

\usepackage[utf8]{inputenc} 
\usepackage[T1]{fontenc}    
\usepackage{lipsum}
\usepackage{natbib}
\usepackage[utf8]{inputenc} 
\usepackage{algorithm}
\usepackage{array}
\usepackage{algpseudocode}
\usepackage{textcomp}
\usepackage{verbatim}
\usepackage{cite}
\usepackage[T1]{fontenc}    
\usepackage{url}            
\usepackage{booktabs}       
\usepackage{amsfonts}       
\usepackage{nicefrac}       
\usepackage{microtype}      
\usepackage{xcolor}         

\usepackage{amsmath}
\usepackage{amsthm}
\usepackage{bm}
\usepackage{color}
\usepackage{graphicx}
\usepackage{makecell}
\usepackage{multirow}
\usepackage{subfigure}
\usepackage{tabularx}
\usepackage{mathtools}
\usepackage{pythonhighlight}
\usepackage{listings}
\usepackage{soul}

\definecolor{pink}{rgb}{0.858, 0.188, 0.478}
\usepackage{xspace}

\newcommand{\figwidth}[4]{
    \begin{figure}[tb]\centering
    \includegraphics[width=#4]{fig/#1}
    \caption{#2}
    \label{#3}\end{figure}
    {}}

\newcommand{\figccwidth}[4]{
    \begin{figure*}[tb]\centering
    \includegraphics[width=#4]{fig/#1}
    \caption{#2}
    \label{#3}\end{figure*}
    {}}

\newcommand{\figtwo}[9]{
    \begin{figure}[tb]\centering
    \subfigure[#2] {
    \label{#3}
    \includegraphics[width=#7]{fig/#1}
    }
    \subfigure[#5] {
    \label{#6}
    \includegraphics[width=#7]{fig/#4}
    }
    \caption{#8}
    \label{#9}
    \end{figure}
    {}}

\newcommand{\rev}[1]{{#1}{}}

\definecolor{commentcolor}{RGB}{110,154,155}   

\usepackage{amsmath,amsfonts,bm}

\def\eqref#1{equation~\ref{#1}}

\def\1{\bm{1}}

\def\vr{{\bm{r}}}

\def\vw{{\bm{w}}}
\def\vx{{\bm{x}}}

\def\vz{{\bm{z}}}

\DeclareMathAlphabet{\mathsfit}{\encodingdefault}{\sfdefault}{m}{sl}
\SetMathAlphabet{\mathsfit}{bold}{\encodingdefault}{\sfdefault}{bx}{n}

\title{SCPL: Enhancing Neural Network Training Throughput with Decoupled Local Losses and Model Parallelism
}

\author{
  Ming-Yao Ho, Cheng-Kai Wang, You-Teng Lin, Hung-Hsuan Chen \\
  Computer Science and Information Engineering \\
  National Central University \\
  Taoyuan, Taiwan\\
  \texttt{horan7854@gmail.com, wang19980531@outlook.com, lyt0310603@gmail.com, hhchen1105@acm.org}
}

\begin{document}
\maketitle

\begin{abstract}

Adopting large-scale AI models in enterprise information systems is often hindered by high training costs and long development cycles, posing a significant managerial challenge. The standard end-to-end backpropagation (BP) algorithm is a primary driver of modern AI, but it is also the source of inefficiency in training deep networks. This paper introduces a new training methodology, Supervised Contrastive Parallel Learning (SCPL), that addresses this issue by decoupling BP and transforming a long gradient flow into multiple short ones. This design enables the simultaneous computation of parameter gradients in different layers, achieving superior model parallelism and enhancing training throughput. Detailed experiments are presented to demonstrate the efficiency and effectiveness of our model compared to BP, Early Exit, GPipe, and Associated Learning (AL), a state-of-the-art method for decoupling backpropagation. By mitigating a fundamental performance bottleneck, SCPL provides a practical pathway for organizations to develop and deploy advanced information systems more cost-effectively and with greater agility. The experimental code is released for reproducibility.\footnote{\url{https://github.com/minyaho/scpl/}}

\end{abstract}


\section{Introduction} \label{sec:intro}

Deep neural networks have become prominent. Although they have excellent predictive power, the practical adoption of these large-scale models in enterprise environments -- for applications ranging from decision support and market forecasting to customer relationship management -- faces a significant challenge: the immense computational cost and time required for training. This challenge is not only technical; it represents a key managerial barrier that hinders the agile development and deployment of advanced information systems, impacting the return on investment of AI initiatives. \rev{Recent industry reports underscore the severity of these financial barriers. For instance, the average cost of computing for enterprise AI is projected to increase by 89\% between 2023 and 2025~\mbox{\citep{ibm2024}}. Furthermore, training costs for frontier AI models have historically increased by 2.4 times annually since 2016~\mbox{\citep{epoch2024}}. In this high-cost landscape, the traditional inefficiency of backpropagation, where sequential dependencies leave expensive hardware underutilized during training, becomes an unsustainable waste of resources.} The scenario where each layer waits for gradient information from the layers closer to the target before proceeding with the gradient computation is called \emph{backward locking}, which severely limits the potential speedup of parallel processing, as the sequential nature of gradient calculations becomes a bottleneck for parallel training.

This paper proposes an innovative methodology, Supervised Contrastive Parallel Learning (SCPL), to address the issue of backward locking. As a result, SCPL realizes \emph{model parallelism}, which not only partitions a large neural network into segments processed across multiple computing units (e.g., GPUs) but also enables parallel processing on these devices. This is similar but different from most of today's model parallelism tools or packages (e.g., the definitions used by Amazon SageMaker\footnote{\url{https://docs.aws.amazon.com/sagemaker/latest/dg/model-parallel-intro.html}}), which allocate different model components to different GPUs but do not necessarily run these GPUs simultaneously. 

SCPL decouples the long gradient flow of a deep neural network. The forward path transforms an input $x$ into the corresponding prediction $\hat{y}$ as a usual neural network in this design. However, the gradient flow on the backward path is blocked between different components (layers). Instead of using a global objective, SCPL assigns a local objective to each component and forces each gradient flow to remain within one component. When allocating these local objectives to different GPUs, SCPL can compute the local gradients without waiting for the gradients in the neighboring layers.

We conduct experiments on multiple open datasets, including natural language processing and computer vision tasks, using famous network structures such as the LSTM, Transformer, vanilla convolutional neural network, VGG, and ResNet. Our results show that SCPL increases training throughput ($1.92\times$ in practice) while maintaining comparable test accuracies compared to backpropagation (BP), Early Exit, GPipe, and Associated Learning (AL)~\citep{wu2022associated, kao2021associated}, a state-of-the-art methodology for decoupling BP. We released the code and Docker images with a step-by-step guide for reproducibility.

The rest of the paper is organized as follows. In Section~\ref{sec:rel-work}, we review previous work on model parallelism and data parallelism. In Section~\ref{sec:method}, we introduce SCPL and its properties. Section~\ref{sec:exp} compares SCPL with BP, Early Exit, GPipe, and AL with respect to their training time and test accuracies. We discuss why SCPL may work, its limitations, and compare and discuss the architectural similarities and differences of SCPL and GPipe in Section~\ref{sec:disc}. We conclude our contribution in Section~\ref{sec:conc}.

\section{Related Work} \label{sec:rel-work}

\subsection{Data Parallelism vs.~Model Parallelism}

Data parallelism (DP) and model parallelism (MP) are two strategies used in distributed deep learning. DP involves distributing batches of training data across multiple devices or processors; each device computes the gradients independently. Devices usually synchronize gradients at the end of every iteration or multiple iterations~\citep{shallue2018measuring, li2020pytorch}. DP is well-suited for scenarios where the model can fit within the memory of each device.

On the other hand, MP mainly tackles the issue of training models that are too large to fit into the memory of a single device. MP divides the model into segments that are processed on different devices. However, the interdependence of the gradient calculations across layers complicates the parallelization process. The layers in one device may need the gradients from other devices to proceed with its gradient calculations, creating a synchronization bottleneck that limits the potential parallelism. As a result, na\"{i}ve model parallelism (which will be called ``NMP'' below) results in poor training efficiency~\citep{huang2019gpipe}.

\subsection{Model Parallelism Strategies}

\begin{table*}[tb]
\caption{A comparison of the properties of the representative related models ($H$: number of layers/components)}
\label{tab:method-adv-cmp}
\centering
\begin{tabular}{@{}cccc@{}}
\toprule
Method & Supervised? & Parallel model training?      & Length of gradient flows \\ \midrule
BP~\citep{rumelhart1986learning} & Y           & N                       & $O(H)$                   \\
AL~\citep{wu2022associated}    & Y           & Y (but not implemented) & $O(1)$                     \\
GPipe~\citep{huang2019gpipe} & Y           & Y                       & $O(H)$                   \\
LoCo~\citep{xiong2020loco} & N           & Y (but not implemented) & $O(1)$                     \\
GIM~\citep{lowe2019putting}   & N           & Y (but not implemented) & $O(1)$                     \\ SCPL (ours) & Y           & Y                       & $O(1)$                     \\
\bottomrule
\end{tabular}
\end{table*}

A core challenge in model parallelism, backward locking, stems from the inherent characteristics of backpropagation (BP). Therefore, the various methods that attempt to find alternative BP methods could be candidates for achieving model parallelism. Studies in this line include target propagation (TP)~\citep{lee2015difference, meulemans2020theoretical, manchev2020target, bengio2014auto}, gradient prediction~\citep{jaderberg2017decoupled}, and local objective assignments~\citep{wu2022associated, kao2021associated, huang2025deinforeg}.  Although many of these methods do not require gradient computation in training, most still need to update the parameters layer-by-layer, so it is challenging to realize model parallelism. Associated learning (AL) is one of the few methods capable of simultaneously updating parameters in different layers. However, the authors of AL released only a sequential implementation, making it difficult to experiment and verify the parallelization capacity of AL.

NMP has a low training efficiency, but MP's training efficiency can be improved by pipelining, which overlaps GPU computations and minimizes idle time between different stages of computation. The most representative model in this line is Google's GPipe~\citep{huang2019gpipe}, which separates mini-batches into micro-batches to pipeline the forward and backward passes. However, GPipe still relies on the chain rule for gradient computation, so bubbles are inevitable, and the length of a gradient flow is $O(H)$, identical to BP. Some of the following works, e.g., PipeDream~\citep{narayanan2019pipedream}, further overlap forward and backward passes, thereby reducing synchronization overhead. However, the lack of global gradient synchronization leads to challenges in maintaining accurate gradient updates, and thus, the final model tends to be underfitting.

Beyond specific pipelining strategies, comprehensive training optimization libraries like DeepSpeed~\citep{rasley2020deepspeed} have emerged. DeepSpeed provides a suite of tools and technologies for large-scale model training to improve scale, speed, and efficiency. It supports various parallelism techniques, including data parallelism, tensor parallelism, and pipeline parallelism (often similar to GPipe, operating on a global loss as discussed in Section~\ref{sec:scpl-vs-gpipe} of some analyses), and critically, memory optimization techniques like the Zero Redundancy Optimizer (ZeRO)~\citep{rajbhandari2020zero}. DeepSpeed primarily optimizes the execution of standard end-to-end backpropagation workflows, rather than altering the fundamental training algorithm itself. The SCPL's approach differs; its core contribution lies in proposing a new training algorithm that inherently facilitates parallelism through local Supervised Contrastive Losses (SCL). The novelty of SCPL is thus centered on its algorithmic design for decoupling and parallelizing the learning process, distinct from providing a general-purpose distributed training execution engine like DeepSpeed. 

Some studies used self-supervised learning to decouple BP, e.g., Greedy InfoMax (GIM)~\citep{lowe2019putting} and LoCo~\citep{xiong2020loco}. Although these studies localize the losses, they mainly focus on modularizing backpropagation and improving model accuracy, but pay little attention to model parallelism. Furthermore, they are self-supervised, not supervised, so their predictive ability in classification or regression is unclear. Although the local losses used in these works could potentially achieve model parallelism via pipelining, their implementations lack support for model parallelism.

Our proposed SCPL is motivated by studies in pipeline model parallelism (e.g., AL and GPipe) and BP decoupling using self-supervised signals (e.g., GIM and LoCo). However, they are different in several ways. First, although GPipe leverages pipelining, it still suffers from backward locking because the gradient flow is identical to BP, and the gradient computation still requires gradients from other layers. For GIM, LoCo, and AL, these methods decouple gradient flows, so the length of each gradient flow is $O(1)$ (that is, its length is irrelevant to the number of layers). As a result, similar to our proposed method, the local losses used by GIM, LoCo, and AL can also be integrated with the pipeline strategy to realize model parallelism. However, these papers' primary focus is not on model parallelism; their released implementations are sequential, requiring significant engineering effort to convert them into versions supporting model parallelism. Moreover, GIM and LoCo primarily target self-supervised rather than fully supervised tasks, so their empirical predictive power on classification/regression is unclear.

A comparison of these models is shown in Table~\ref{tab:method-adv-cmp}. 

\section{Methodology} \label{sec:method}

This section details the SCPL methodology, designed to enhance the throughput of neural network training. We begin in Section~\ref{sec:scpl-foundation} by laying the conceptual groundwork, explaining the prevailing training challenges that motivate SCPL, and introducing its core principles. Following this, Section~\ref{sec:scpl-prelim} reviews the essential preliminaries, specifically focusing on contrastive learning and the Supervised Contrastive Learning (SCL) framework that forms a cornerstone of our approach. Section~\ref{sec:scpl-core-method} elaborates on the central mechanism of SCPL: how we decouple end-to-end backpropagation by strategically employing local SCL objectives within network segments and how these segments are efficiently trained using pipelined model parallelism. Subsequently, Section~\ref{sec:forward-backward-inf-paths} details the forward path, the localized backward paths, and the distinct inference function within the SCPL architecture. Section~\ref{sec:scpl-pipeline} provides implementation details and an illustrative pseudocode to clarify the operational aspects of SCPL in integration with pipelining, with a pseudocode given in Section~\ref{sec:pseudo}.

This section details the operational mechanisms of SCPL. Although the throughput gains derived from pipelined parallelism are readily apparent from the architectural design, the factors contributing to its high prediction accuracy -- specifically regarding the use of local objectives -- are less immediately intuitive. We therefore dedicate Section~\ref{sec:why-scpl-works} to analyzing why this approach yields effective learning.

\subsection{Conceptual Foundations for SCPL: Addressing Training Bottlenecks} \label{sec:scpl-foundation}

Training modern deep neural networks is largely based on the backpropagation algorithm, which is computationally demanding. In intuitive terms, backpropagation enables a model to learn from its predictive errors by iteratively adjusting its internal parameters, working backward from the output layer through its many hidden layers. However, this inherent sequential dependency, where adjustments in one layer often depend on computations from preceding layers in the backward path, can introduce significant inefficiencies, especially as networks grow deeper. This challenge, sometimes referred to as \emph{backward locking}~\citep{wu2022associated}, can limit the effective utilization of parallel processing, and thus restrict the overall training throughput.

To mitigate these training bottlenecks, SCPL proposes an alternative training paradigm centered on decoupling the conventional end-to-end backpropagation. This strategy is similar to decomposing a single, extensive task into a series of shorter, more manageable subtasks that can be executed in parallel. SCPL achieves this by integrating ``local losses'' at designated intermediate points within the network. These local losses function as intermediate targets, forcing each segment to independently learn discriminative representations. SCPL distinctively employs Supervised Contrastive Learning (SCL) for these local objectives. The SCL guides each network segment to develop representations where inputs belonging to the same class are drawn closer in the embedding space, while those from different classes are pushed further apart.

The training of these segments, each with its local SCL objective, is orchestrated via pipelined parallelism. This mechanism operates much like an industrial assembly line: as one mini-batch of data completes its local forward (and local backward) passes in an early segment of the network, it proceeds to the next segment, while the initial segment simultaneously begins processing the subsequent mini-batch. This overlapping of computation across different segments and mini-batches significantly enhances hardware utilization. Through this synergistic combination of decoupled local losses (specifically SCL) and pipelined model parallelism, SCPL aims to improve training throughput, thereby accelerating the training phase of deep neural networks while endeavoring to preserve or even enhance the predictive accuracy achieved by traditional end-to-end training approaches.

\subsection{Preliminaries: contrastive learning and supervised contrastive learning} \label{sec:scpl-prelim}

\figccwidth{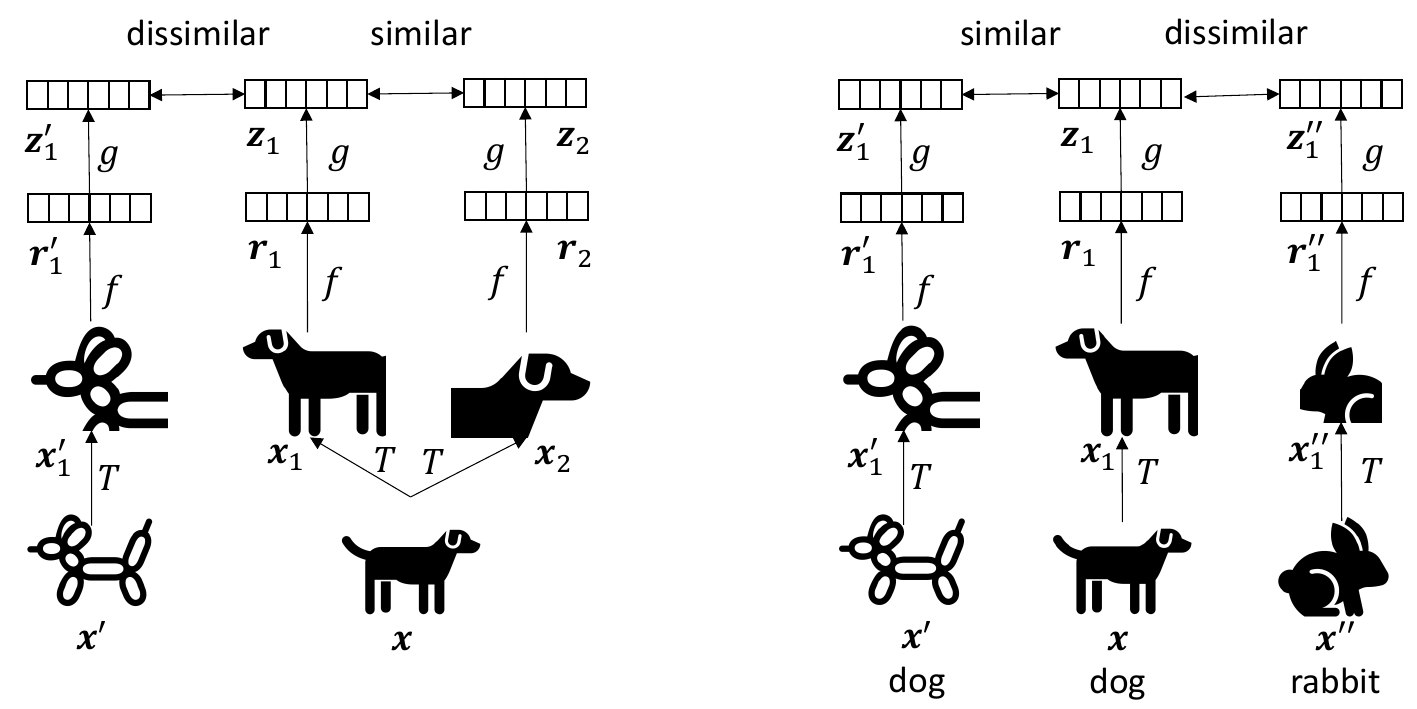}{An illustration of contrastive learning (CL) and supervised contrastive learning (SCL). CL regards an anchor image's augmented images as positive pairs (e.g., $\boldsymbol{x}_1$ and $\boldsymbol{x}_2$ above) and regards the anchor image's augmented image to all non-augmented images as negative pairs (e.g., $\boldsymbol{x}_1$ and $\boldsymbol{x}_1'$ on the left). SCL regards augmented images as positive pairs if they have the same label (e.g., $\boldsymbol{x}_1$ and $\boldsymbol{x}_1'$ on the right); a pair of augmented images is a negative pair if their labels are different (e.g., $\boldsymbol{x}_1$ and $\boldsymbol{x}_1''$).}{fig:cl-vs-scl}{4.6in}

Contrastive learning (CL) is a self-supervised technique for learning the representations of objects. Referring to the left of Figure~\ref{fig:cl-vs-scl}, given an image $\vx$, CL involves generating different views (i.e., $\vx_1$ and $\vx_2$) through the same family of data augmentations $T$. The generated views ($\vx_1$ and $\vx_2$) are further transformed using an encoder $f$ and a projection head $g$ to minimize contrastive loss between the output vectors (i.e., $\vz_1$ and $\vz_2$). After training, the projection head $g$ is disregarded, and only the encoder $f$ is used to generate the representations of the images~\citep{chen2020simple}. In other words, given an anchor image $\vx$, CL regards $\vx$'s augmented images as positive instances and all other images as negative instances.
Positive pairs should be similar after encoding and projection.

Supervised Contrastive Learning (SCL) extends CL from a self-supervised setting to a fully supervised setting. Therefore, the training dataset for SCL consists of both the training features and the labels. Referring to the right of Figure~\ref{fig:cl-vs-scl}, given an anchor image $\vx$ with label $c$, the positive instances include the augmented images of $\vx$ and other images (along with their augmented images) of label $c$ in the same batch, while all other images in the same batch are considered negative instances~\citep{khosla2020supervised}.

\subsection{Decoupling end-to-end backpropagation via supervised contrastive learning} \label{sec:scpl-core-method}

\figccwidth{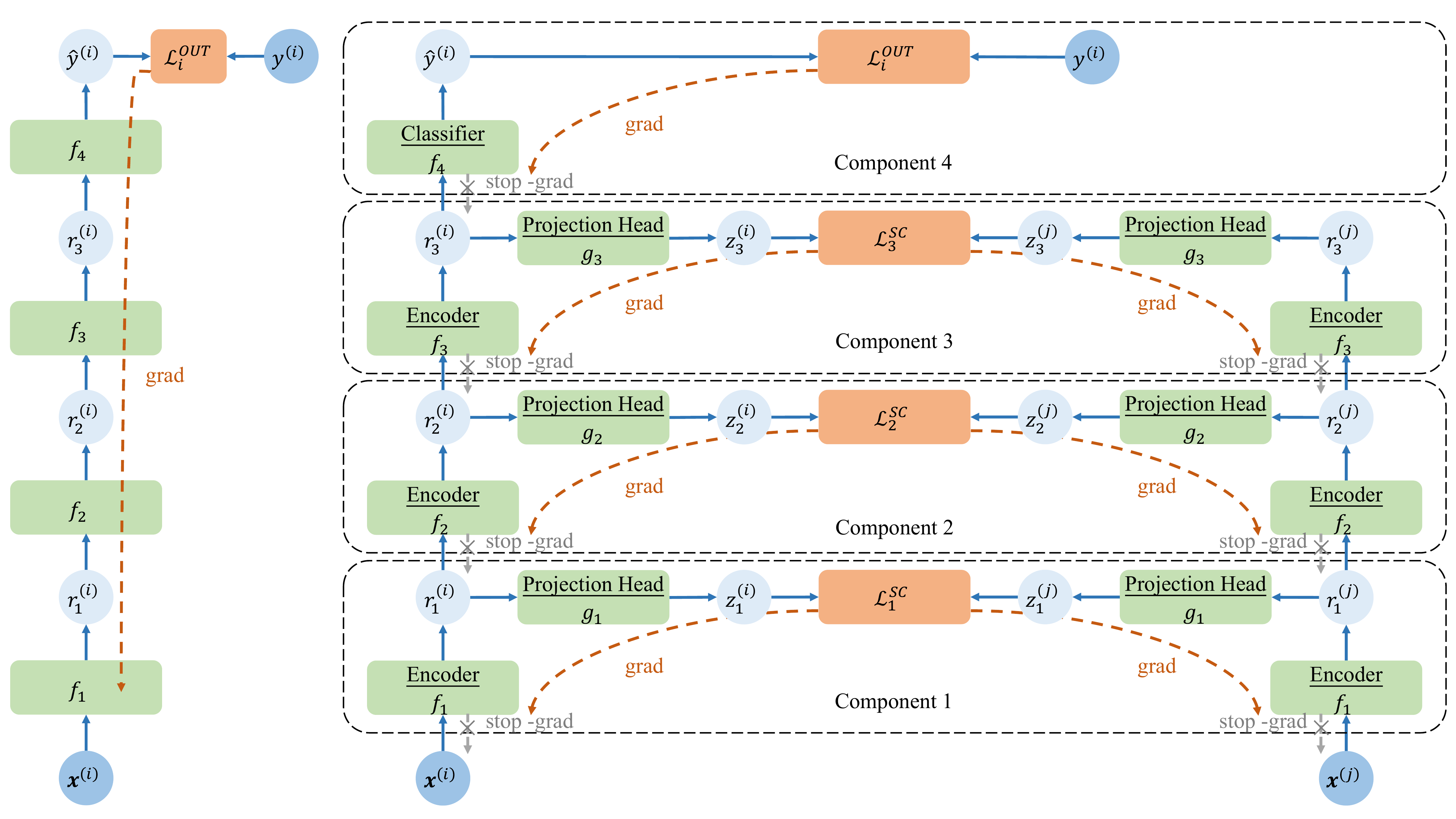}{An example neural network with 3 hidden layers (left) and its corresponding SCPL network (right). Solid blue arrows correspond to forward paths, red dashed arrows correspond to backward paths, green boxes denote the parameters (functions), and orange boxes represent the loss functions. 
Gradient flows are blocked between neighboring blocks for SCPL.}{fig:nn-vs-scpl}{\textwidth}

This section presents SCPL, which leverages supervised contrastive learning to split one long gradient flow in a deep neural network into multiple shorter ones.

As an example, let us first consider a standard neural network with 3 hidden layers. As shown on the left of Figure~\ref{fig:nn-vs-scpl}, $\vx^{(i)}$ refers to an input image $i$, and the function $f_{\ell}~(\ell=1,\ldots,4)$ transforms $\vr_{\ell-1}^{(i)}$ into $\vr_\ell^{(i)}$ (under the assumption that $\vx^{(i)} = \vr_0^{(i)}$ and the predicted class $\hat{y}^{(i)} = \vr_4^{(i)}$). Depending on the network architecture, the functions $f_\ell$ could be various neural network layers, such as fully connected layers, convolutional layers, pooling layers, or residual blocks. The objective $\mathcal{L}^{OUT}$ is determined by the task type. For example, a classification task typically uses the cross-entropy loss between the predicted $\hat{y}^{(i)}$ and the ground-truth class $y^{(i)}$ as $\mathcal{L}^{OUT}$. We use backpropagation to obtain $\partial \mathcal{L}^{OUT} / \partial \theta_{f_\ell}$ for each layer $\ell$, where $\theta_{f_\ell}$ represents the parameters of the function $f_\ell$. Once the gradients are obtained, we can use gradient-based optimization strategies, e.g., gradient descent, to update the parameter values. Given a neural network with $H$ hidden layers, the longest gradient flow is constructed as a product of $H+2$ local gradients. For example, to obtain $\partial \mathcal{L}^{OUT} / \theta_{f_1}$ in a network with 3 hidden layers (as shown on the left of Figure~\ref{fig:nn-vs-scpl}), we need the following:

\begin{equation} \label{eq:chain-rule-bp}
    \frac{\partial \mathcal{L}^{OUT}}{\partial \theta_{f_1}} = \frac{\partial \mathcal{L}^{OUT}}{\partial \hat{y}^{(i)}} \times \frac{\partial \hat{y}^{(i)}}{\partial r_3^{(i)}} \times \frac{\partial r_3^{(i)}}{\partial r_2^{(i)}} \times \frac{\partial r_2^{(i)}}{\partial r_1^{(i)}} \times \frac{\partial r_1^{(i)}}{\partial \theta_{f_1}}.
\end{equation}

The number of terms in this product grows linearly with the depth of the network. Therefore, as a network becomes deeper, its long gradient flow may cause optimization and performance issues, as discussed in Section~\ref{sec:intro}.

We use Figure~\ref{fig:nn-vs-scpl} to illustrate our strategy of cutting a long gradient flow into several local gradients for a neural network with 3 hidden layers. Let $\vr_0^{(i)}$ (i.e., $\vx^{(i)}$) and $\vr_0^{(j)}$ (i.e., $\vx^{(j)}$) be two image views (i.e., augmented images) in the same batch ($\vr_0^{(i)}$ and $\vr_0^{(j)}$ may or may not be augmented images of the same image). We use $f_1$ to transform each of them, obtaining $\vr_1^{(i)}$ and $\vr_1^{(j)}$, and further use the function $g_1$ to convert them into $\vz_1^{(i)}$ and $\vz_1^{(j)}$, respectively. The functions $f_1$ and $g_1$ can be considered as the encoder and projection head, respectively, in SCL (refer to Figure~\ref{fig:cl-vs-scl}). We repeat the same process for each hidden layer $\ell$ to form the corresponding component $\ell$. If $\vx^{(i)}$ and $\vx^{(j)}$ are two different views of the same image or if $y^{(i)}$ (the label of $\vx^{(i)}$) is the same as $y^{(j)}$ (the label of $\vx^{(j)}$), then we should ensure that $\vz_{\ell}^{(i)}$ is close to $\vz_{\ell}^{(j)}$ for all $\ell$. Otherwise, we should increase their distance. Eventually, we define the local supervised contrastive loss $L^{SC}_{\ell}$ for the batch $B$ in the layer $\ell$ as below.

\begin{equation} \label{eq:scl-loss}
    \mathcal{L}_{\ell}^{SC}(B) =  \sum_{\forall i \in B} \frac{-1}{|P(i)|}\sum_{\forall p \in P(i)} \log \frac{\exp\left(\vz_{\ell}^{(i)} \cdot \vz_{\ell}^{(p)} / \tau \right)}{\sum_{\substack{\forall j \in B \\ j\neq i}} \exp\left(\vz_{\ell}^{(i)} \cdot \vz_{\ell}^{(j)} / \tau \right)},
\end{equation}
where $B = \{1, 2, \ldots, b\}$ represents a batch of multiview images ($b=2N$ if using data augmentation and $b=N$ otherwise), $P(i)$ is the set of all positive samples for an image $i$, and $\tau$ is a hyperparameter.

Ultimately, the global objective function of a batch $B$ is an accumulation of local supervised contrastive losses and losses in the output layer (i.e., the distance between $\hat{y}^{(i)}$ and $y^{(i)}$):

\begin{equation} \label{eq:global-loss}
    \mathcal{L}(B) = \sum_{\ell=1}^H \mathcal{L}_{\ell}^{SC}(B) + \sum_{\forall i \in B} \mathcal{L}_i^{OUT},
\end{equation}
where $H$ is the number of hidden layers, and $\mathcal{L}_i^{OUT}$ is the $i$th loss in the output layer (refer to the right of Figure~\ref{fig:nn-vs-scpl}).

Detailed structures and hyperparameters are given in the released code. To help understand, the computation of $\mathcal{L}_{\ell}^{SC}$ and the pseudocode of SCPL for a 3-layer vanilla ConvNet are given in Algorithm~\ref{alg:supconloss} and Algorithm~\ref{alg:scpl-no-pipeline}, respectively, in Section~\ref{sec:pseudo}.

\subsection{Forward path, backward path, and inference function} \label{sec:forward-backward-inf-paths}

For a regular neural network (e.g., left of Figure~\ref{fig:nn-vs-scpl}), the forward path and the inference function are identical, and the backward path is obtained by inverting the direction of the forward path. However, SCPL is different because we divide the objective into several local ones. Consequently, we have multiple short local forward paths, multiple short local backward paths, and one global inference path; the inference path and the forward paths are no longer identical in SCPL because some paths are only used in training.

Each component $\ell$ has its forward and backward paths. Taking the SCPL network in Figure~\ref{fig:nn-vs-scpl} as an example, the forward path of the component $\ell$ transforms each $\vr_{\ell-1}^{(i)}$ into $\vr_{\ell}^{(i)}$ through the local encoder $f_{\ell}$ and further transforms each $\vr_{\ell}^{(i)}$ into $\vz_{\ell}^{(i)}$ via the local projection head $g_{\ell}$. In the backward path, each hidden layer computes $\partial \mathcal{L}^{SC}_{\ell} /\partial \theta_{g_{\ell}}$ and $\partial \mathcal{L}^{SC}_{\ell} /\partial \theta_{f_{\ell}}$ based on the chain rule and updates the parameters using gradient-based optimization strategies. We block the gradient flow between each component. As a result, each gradient flow remains within one component. Equation~\ref{eq:chain-rule-cl-g} and Equation~\ref{eq:chain-rule-cl-f} show these local gradient flows.

\begin{equation} \label{eq:chain-rule-cl-g}
    \frac{\partial \mathcal{L}_{\ell}^{SC}}{\partial \theta_{g_{\ell}}} = 
    \frac{\partial \mathcal{L}_{\ell}^{SC}}{\partial \vz_{\ell}^{(i)}} \times  \frac{\partial \vz_{\ell}^{(i)}}{\partial \theta_{g_{\ell}}}.
\end{equation}

\begin{equation} \label{eq:chain-rule-cl-f}
    \frac{\partial \mathcal{L}_{\ell}^{SC}}{\partial \theta_{f_{\ell}}} = 
    \frac{\partial \mathcal{L}_{\ell}^{SC}}{\partial \vz_{\ell}^{(i)}} \times  \frac{\partial \vz_{\ell}^{(i)}}{\partial \vr_{\ell}^{(i)}} \times \frac{\partial \vr_{\ell}^{(i)}}{\partial \theta_{f_{\ell}}}.
\end{equation}

Eventually, even if we construct a deep neural network, the cost of computing each $\partial \mathcal{L}^{SC} /\partial \theta_{f_{\ell}}$ and each $\partial \mathcal{L}^{SC} /\partial \theta_{g_{\ell}}$ remains constant (i.e., $O(1)$). Additionally, the gradient flow in the output layer is also short: we compute $\partial \mathcal{L}_k^{out} / \partial \theta_{f_{H+1}}$ (where $H$ is the number of hidden layers).  Consequently, this design may alleviate various issues caused by long gradient flows.

In the inference (prediction) phase, we need only the encoders $f_{\ell}$ but not the projection heads $g_{\ell}$, as shown by Equation~\ref{eq:inf-func}:

\begin{equation} \label{eq:inf-func}
    \hat{y}^{(i)} = f_{H+1} \circ f_H \circ \ldots \circ f_2 \circ f_1 (\vx^{(i)}),
\end{equation}
where $\circ$ is the function composition operator ($H=3$ for the example illustrated in Figure~\ref{fig:nn-vs-scpl}).

Although our proposed method (e.g., right of Figure~\ref{fig:nn-vs-scpl}) involves more parameters than a standard neural network structure (e.g., left of Figure~\ref{fig:nn-vs-scpl}) during training, they have the same number of parameters during inference because both of them use only the functions $f_{\ell}$. Therefore, they have the same hypothesis space. The parameters that participate in the inference phase ($\theta_{f_{\ell}}$-s) are called the \emph{effective parameters}. The parameters used during training but not during inference ($\theta_{g_{\ell}}$-s) are called the \emph{affiliated parameters}. As a bonus effect, having affiliated (redundant) parameters sometimes helps optimization~\citep{arora2018optimization, chen2020accelerating}.

\subsection{Model parallelism via local objectives and pipelining} \label{sec:scpl-pipeline}

\figccwidth{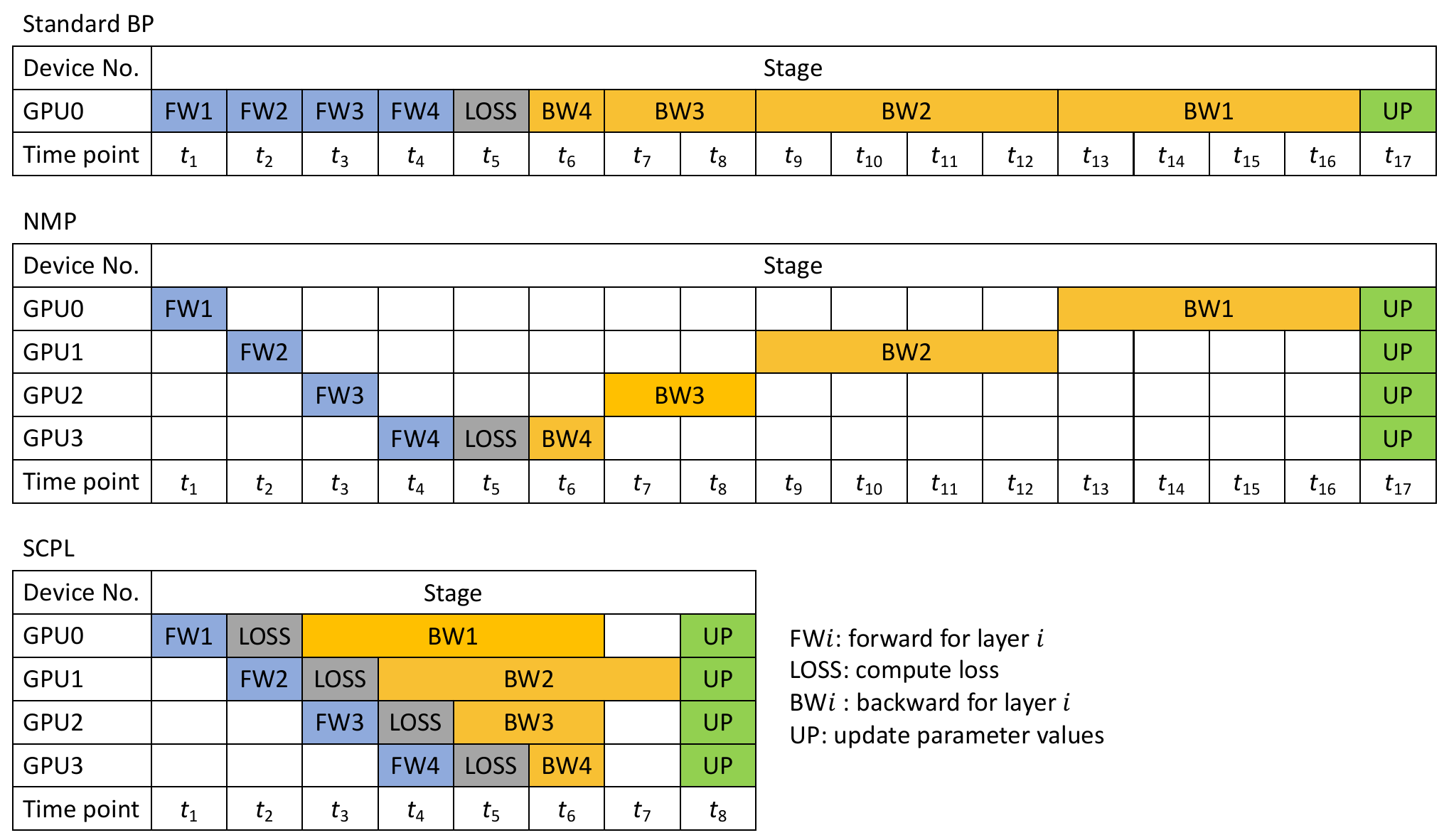}{An illustrating example to compare the GPU usage of one iteration for standard BP, NMP, and SCPL. The GPU utilization in a real environment is shown in Figure~\ref{fig:lstm-imdb-profiler-cmp}.}{fig:bp-vs-scpl}{5.8in}

Since each component has its local objective, we can parallelize the training procedure by pipelining. We use the network illustrated in Figure~\ref{fig:nn-vs-scpl} as an example. Referring to Figure~\ref{fig:bp-vs-scpl}, the top subfigure shows a standard learning iteration using a single GPU. When a model is too large to fit in a single GPU, we have to resort to NMP, as illustrated in the middle subfigure, which segments the model into 4 components and assigns each part to one GPU (we ignore the communication cost between GPUs in the illustration). However, the GPUs cannot operate simultaneously due to the dependencies between different components, so bubbles exist. Finally, the lower subfigure illustrates SCPL: in time unit $t_1$, the 1st GPU (GPU0) computes the forward path in component 1. At $t_2$, the 2nd GPU takes $\vr_1^{(i)}$ and $\vr_1^{(j)}$ as input to perform forward, and the 1st GPU computes the local loss for component 1 at the same moment.  At $t_3$, the 3rd GPU takes $\vr_2^{(i)}$ and $\vr_2^{(j)}$ as input to perform forward, the 2nd GPU computes the local loss for component 2, and the 1st GPU computes the gradients for the parameters in component 1 via backpropagation. Therefore, different GPUs may compute the parameter gradients in different components simultaneously~\citep{huang2025deinforeg}.

Referring to Figure~\ref{fig:lstm-imdb-profiler-cmp} in Section~\ref{sec:profiler}, we use a PyTorch profiler to show the utilization of GPUs and a CPU in a real training job. The profiler shows the GPU usage footprints of NMP and SCPL, which are indeed close to our illustration in Figure~\ref{fig:bp-vs-scpl}.


\subsection{Pseudo code} \label{sec:pseudo}

To help understand the details of SCPL, we provide pseudocodes for local supervised contrastive losses (Algorithm~\ref{alg:supconloss}) and SCPL without pipelining (Algorithm~\ref{alg:scpl-no-pipeline}) of a 3-layer vanilla ConvNet.

\definecolor{codegreen}{rgb}{0,0.6,0}
\definecolor{codegray}{rgb}{0.5,0.5,0.5}
\definecolor{codepurple}{rgb}{0.58,0,0.82}
\definecolor{backcolour}{rgb}{0.95,0.95,0.92}

\lstdefinestyle{mystyle}{
  backgroundcolor=\color{backcolour}, commentstyle=\color{codegreen},
  keywordstyle=\color{magenta},
  numberstyle=\tiny\color{codegray},
  stringstyle=\color{codepurple},
  basicstyle=\ttfamily\footnotesize,
  breakatwhitespace=false,         
  breaklines=true,                 
  captionpos=b,                    
  keepspaces=true,                 
  numbers=left,                    
  numbersep=5pt,                  
  showspaces=false,                
  showstringspaces=false,
  showtabs=false,                  
  tabsize=2
}

\lstset{style=mystyle}
\renewcommand{\lstlistingname}{Algorithm}
\renewcommand{\lstlistlistingname}{List of \lstlistingname s}

\begin{figure}[tbh]
\begin{lstlisting}[language=Python, caption={PyTorch-like pseudocode for $L^{sc}$}, label={alg:supconloss}]
import torch
import torch.nn as nn

class SupConLoss(nn.Module):
    def __init__(self, dim):
        super.__init__()
        self.linear = nn.Sequential(nn.Linear(dim, 512), nn.ReLU(), nn.Linear(512, 1024))
        self.temperature = 0.1
        
    def forward(self, x, label):
        x = self.linear(x)
        x = nn.functional.normalize(x)
        label = label.view(-1, 1)
        bsz = label.shape[0]
        mask = torch.eq(label, label.T).float()
        anchor_mask = torch.scatter(torch.ones_like(mask), 1, torch.arange(bsz).view(-1, 1), 0)
        logits = torch.div(torch.mm(x, x.T), self.temperature) deno = torch.exp(logits) * anchor_mask
        prob = logits - torch.log(deno.sum(1, keepdim=True)) 
        loss = -(anchor_mask * mask * prob).sum(1) / mask.sum() 
        return loss.view(1, bsz).mean()
\end{lstlisting}
\end{figure}

\begin{figure}[tbh]
\begin{lstlisting}[language=Python, caption={PyTorch-like pseudocode for SCPL without pipelining}, label={alg:scpl-no-pipeline}]
import torch
import torch.nn as nn

# A simple 3-layer CNN example for SCPL architecture.
class CNN_SCPL(nn.Module):
    def __init__(self, dim): 
        super.__init__()
        CNNs = [ ]
        losses = [ ]
        channels = [3, 128, 256, 512] self.shape = 32
        for i in range(3):
            CNNs.append(nn.Sequential(nn.Conv2d(channels[i], channels[i+1], padding=1), nn.ReLU())
            losses.append(SupConLoss(self.shape*self.shape*channels[i+1]))
        self.CNN = nn.ModuleList(CNNs)
        self.loss = nn.ModuleList(losses)
        self.fc = nn.Sequential(flatten(), nn.Linear(self.shape*self.shape*channels[-1], 10))
        self.ce = nn.CrossEntropyLoss()

    def forward(self, x, label): loss = 0
        for i in range(3):
            # .detach() prevents a gradient flows to the neighboring layer
            x = self.CNN[i](x.detach()) 
            if self.training:
                loss += self.loss[i](x, label)
        y = self.fc(x.detach())
        if self.training:
            loss += self.ce(y, label)
            return loss
        return y
\end{lstlisting}
\end{figure}

\section{Experiments} \label{sec:exp}

We compare SCPL with baselines using different neural networks on text datasets (AG's news and IMDB) and image datasets (CIFAR-10, CIFAR-100, and Tiny-ImageNet). Baseline models include BP, the Early Exit mechanism (to be introduced later), GPipe, and AL, a state-of-the-art method for BP decoupling in terms of test accuracy. We test LSTM and Transformer for text datasets. We test the VGG network and the residual network (ResNet) for the image datasets. Detailed hyperparameter settings can be found in our released code.

\subsection{Speedup of the empirical training time} \label{sec:training-time-cmp}



\begin{table*}[tb]
\caption{The speedup of the training time for SCPL (1, 2, or 4 GPUs) and GPipe (1, 2, or 4 GPUs) using BP of the same batch size as the reference.  The actual running minutes of BP are shown in parentheses. We use the Transformer as the backbone network and the IMDB as the dataset.}
\label{tab:speedup-cmp-transformer-imdb}
\centering
\begin{tabular}{@{}c|ccccc@{}}
\toprule
Batch size    & 32            & 64            & 128           & 256           & 512 \\ \midrule
BP            & 1x (196 min)  & 1x (173 min)  & 1x (156 min)  & 1x (149 min)  & 1x (147 min)  \\
GPipe (1 GPU) & 0.75x & 0.72x & 0.72x & 0.71x & 0.70x  \\
GPipe (2 GPUs) & 1.00x & 0.92x & 0.93x & 0.93x & 0.92x \\
GPipe (4 GPUs) & 1.35x & 1.25x & 1.17x & 1.16x & 1.11x \\
SCPL (1 GPU)  & 1.12x & 1.07x & 1.03x & 1.03x & 1.05x \\
SCPL (2 GPUs) & 1.43x & 1.37x & 1.32x & 1.37x & 1.38x  \\
SCPL (4 GPUs) & 1.92x & 1.82x & 1.66x & 1.67x & 1.66x \\ \bottomrule
\end{tabular}
\end{table*}

\begin{table*}
\caption{The speedup of the training time for SCPL (1, 2, or 4 GPUs) and GPipe (1, 2, or 4 GPUs) using BP of the same batch size as the reference.  The actual running minutes of BP are shown in parentheses. We use VGG as the backbone network and Tiny-ImageNet as the dataset.}
\label{tab:speedup-cmp-vgg-tinyimgnet}
\centering
\begin{tabular}{@{}c|ccccc@{}}
\toprule
Batch size    & 32            & 64            & 128           & 256           & 512   \\ \midrule
BP            & 1x (204 min)  & 1x (215 min)  & 1x (220 min)  & 1x (224 min)  & 1x (244 min) \\
GPipe (1 GPU) & 0.45x & 0.54x & 0.62x & 0.63x & 0.67x \\
GPipe (2 GPUs) & 0.57x & 0.66x & 0.67x & 0.82x & 0.84x \\
GPipe (4 GPUs) & 0.73x & 0.92x & 1.00x & 1.05x & 1.27x \\
SCPL (1 GPU)  & 0.66x & 0.80x & 0.89x & 0.92x & 0.98x \\
SCPL (2 GPUs) & 0.82x & 0.98x & 0.97x & 1.19x & 1.24x \\
SCPL (4 GPUs) & 1.04x & 1.37x & 1.44x & 1.53x & 1.92x \\ \bottomrule
\end{tabular}
\end{table*}


We compare the training time of SCPL (with 1, 2, or 4 GPUs) with BP and GPipe (with 1, 2, or 4 GPUs), a representative MP method. The speedup of a method $m$ is defined as the practical training time of BP divided by the training time of $m$. We do not compare the training time with other methods that employ local losses (e.g., GIM, LoCo, or AL) for several reasons. First, these methods primarily focus on modularizing backpropagation, and model parallelism is not their primary objective. Consequently, their design and optimization criteria differ from those of SCPL, making direct comparisons less meaningful. Second, the publicly available codes for these methods do not include implementations of model parallelism. Thus, adapting their frameworks for a fair comparison with SCPL is challenging. Finally, even if modifications were possible, the resulting implementations might not be optimized or representative of the methods' true capabilities. We do compare SCPL's training time with GPipe under different GPU numbers.

Table~\ref{tab:speedup-cmp-transformer-imdb} shows the speedup of training time under different batch sizes for SCPL and GPipe (with 1, 2, or 4 GPUs) when using Transformer as the network architecture and IMDB as the experimental datasets. Similarly, using VGG as the network architecture and Tiny-ImageNet as the experimental dataset, Table~\ref{tab:speedup-cmp-vgg-tinyimgnet}
compares the speedup of training time. We use Transformer and VGG because they are representative network structures for natural language processing tasks and vision tasks. 

Here are our observations. First, SCPL indeed accelerates the training time as we have more GPUs. However, due to the communication and synchronization overheads and load balancing issues, the speedup improves sub-linearly with the number of GPUs. Second, although the training efficiency of GPipe improves when we use more GPUs, the improved ratio is worse than that of SCPL. This is likely because GPipe still suffers from the issue of backward locking, so the bubbles are unavoidable. Additionally, although the original GPipe paper~\citep{huang2019gpipe} reported a speedup from 1.7x to 1.8x when using 4 GPUs, these speedups were obtained by setting the number of micro-batches to an extreme number (32). Later experiments have revealed that when using GPipe with multiple GPUs, the training time is generally longer than traditional BP with a single GPU and NMP~\citep{zhang2023experimental}. This observation is consistent with our experimental results. Third, despite involving more parameters during training, SCPL with a single GPU has a training time similar to, and sometimes faster than, BP. It may initially seem counterintuitive because SCPL involves more operations, and a single GPU does not seem to parallelize the computation loading. However, executing various computation parts on a GPU is asynchronous, so a larger GPU can potentially run multiple kernels simultaneously. Due to the asynchronous property, decoupling the computation means that the update of the first component can initiate while the forward propagation of the remaining network occurs, potentially leading to improved GPU utilization. However, this favorable scenario is not always guaranteed and relies heavily on factors such as tensor sizes and GPU specifications, which might explain why it is not consistently observed. That being said, this exciting discovery makes SCPL an attractive option even in a single GPU environment.


All experiments are tested on a Container Compute Service with NVIDIA Tesla V100 GPUs.

\subsection{Accuracy comparison}

\begin{table}[tb]
\caption{A comparison of the test accuracies (mean $\pm$ standard deviation) of different methodologies when using different neural network architectures on IMDB.  We highlight the winner among the non-BP methodologies and all models that are non-significantly different from the best models in boldface. We mark a $\dag$ symbol if the test accuracy of this methodology is higher than that of BP.}
\label{tab:imdb-cmp}
\centering
\begin{tabular}{c||cc}
\toprule
 & LSTM & Transformer \\ 
\midrule
BP & $89.68 \pm 0.20$ & $87.54 \pm 0.44$ \\
\midrule
Early Exit & $84.34 \pm 0.31$ & $80.24 \pm 0.24$ \\
AL & $86.41 \pm 0.61$ & $ 85.65 \pm 0.77$ \\
SCPL & $\boldsymbol{89.84} \pm 0.10~\dag$ & $\boldsymbol{89.03} \pm 0.12~\dag$ \\
\bottomrule
\end{tabular}
\end{table}

\begin{table}[tb]
\caption{A comparison of the test accuracies of different methodologies when using different neural network architectures on AG's news. We follow the same notations used in Table~\ref{tab:imdb-cmp}.}
\label{tab:agnews-cmp}
\centering
\begin{tabular}{c||cc}
\toprule
 & LSTM & Transformer \\ 
\midrule
BP & $91.97 \pm 0.19$ & $91.27 \pm 0.18$ \\
\midrule
Early Exit & $85.91 \pm 0.11$ & $85.79 \pm 0.43$ \\
AL & $91.53 \pm 0.20$ & $ \boldsymbol{91.17} \pm 0.43$ \\
SCPL & $\boldsymbol{92.12} \pm 0.04~\dag$ & $\boldsymbol{91.64} \pm 0.23~\dag $\\
\bottomrule
\end{tabular}
\end{table}

\begin{table}[tb]
\caption{A comparison of the test accuracies of different methodologies when using different neural network architectures on Tiny-ImageNet. We follow the same notations used in Table~\ref{tab:imdb-cmp}.}
\label{tab:tiny-imagenet-cmp}
\centering
\begin{tabular}{c||cc}
\toprule
 & VGG & ResNet \\ 
\midrule
BP & $48.30 \pm 0.14$ & $49.71 \pm 0.18$ \\
\midrule
Early Exit & $46 \pm 0.18$ & $ 40 \pm 0.34$ \\
AL & $\boldsymbol{49.06} \pm 0.14~\dag$ & $44.83 \pm 0.15$ \\
SCPL & $\boldsymbol{48.95} \pm 0.17~\dag$  & $\boldsymbol{46.87} \pm 0.26$ \\
\bottomrule
\end{tabular}
\end{table}

This section reports comparisons of the test accuracies of models trained by BP, the Early Exit mechanism, and AL, a state-of-the-art method for BP decoupling in terms of test accuracy. Early Exit refers to the strategy of assigning a local objective to a component by adding a local auxiliary classifier that outputs a predicted $\hat{y}$ and updating local parameters based on the difference between $\hat{y}$ and the ground truth target $y$. We omit GPipe because its accuracy would theoretically be the same as BP. We extensively search for the appropriate hyperparameters for each model to ensure fair comparisons. All models are trained for 200 epochs. We repeat each experiment five times and report the average and standard deviation in all the following tables. 

\begin{table}[tb] 
\caption{A comparison of the test accuracies of different methodologies when using different neural network architectures on CIFAR-10.  We follow the same notations used in Table~\ref{tab:imdb-cmp}.}
\label{tab:cifar10-cmp}
\centering
\begin{tabular}{c||ccc}
\toprule
           & Vanilla ConvNet & VGG        & ResNet     \\ 
\midrule
BP         & $86.85 \pm 0.57$      & $93.02 \pm 0.03$ & $93.95 \pm 0.11$ \\
\midrule
Early Exit & $83.16 \pm 0.33$      & $91.28 \pm 0.15$ & $89.63 \pm 0.34$ \\
AL         & $\boldsymbol{86.98} \pm 0.24~\dag$   & $\boldsymbol{93.22} \pm 0.12~\dag$ & $91.33 \pm 0.09$ \\
SCPL   & $\boldsymbol{86.98} \pm 0.33~\dag$ & $\boldsymbol{93.42} \pm 0.11~\dag$ & $\boldsymbol{92.78} \pm 0.11$ \\ 
\bottomrule
\end{tabular}
\end{table}

\begin{table}[tb]
\caption{A comparison of the test accuracies of different methodologies when using different neural network architectures on CIFAR-100.  We follow the same notations used in Table~\ref{tab:imdb-cmp}.}
\label{tab:cifar100-cmp}
\centering
\begin{tabular}{c||ccc}
\toprule
 & Vanilla ConvNet & VGG & ResNet \\ 
\midrule
BP & $58.68 \pm 0.13$ & $72.58 \pm 0.39$ & $73.59 \pm 0.11$ \\
\midrule
Early Exit & $ 50.64 \pm 0.44$ & $ 71.11 \pm 0.95$ & $ 64.48 \pm 0.41 $ \\
AL & $53.06 \pm 0.15$ & $72.43 \pm 0.27$ & $67.53 \pm 0.32$ \\
SCPL & $\boldsymbol{59.63} \pm 0.37~\dag$ & $\boldsymbol{73.14} \pm 0.30~\dag$  & $\boldsymbol{70.41} \pm 0.27$ \\
\bottomrule
\end{tabular}
\end{table}

Table~\ref{tab:imdb-cmp} shows the results on the IMDB, a text classification dataset. We use pre-trained Glove word embeddings~\citep{pennington2014glove} of dimensionality 300 in the first layer of the model for both LSTM and Transformer. The simple Early Exit mechanism can be used to learn the relationship between the texts and the corresponding class. However, the test accuracies of Early Exit are much worse than those of BP. This is likely because Early Exit loses too much information about the input by reducing the representation to fit the labels in each layer. Such greedy behavior may obtain less-optimal representations. AL yields test accuracies better than Early Exit on both LSTM and Transformer. SCPL performs best among methodologies that use local objectives for training. The test accuracies outperform BP. The experimental results on AG's news, another text classification task, are presented in Table~\ref{tab:agnews-cmp}.
The results are similar: Early Exit is the worst, and SCPL produces the best among local objective-based learning strategies and outperforms BP on both the LSTM and Transformer architectures.

Table~\ref{tab:tiny-imagenet-cmp} gives the results on an image classification dataset, Tiny-ImageNet. AL and our proposed SCPL yield test accuracies that are better than those of BP based on the VGG architectures. However, when ResNet is used, BP yields the highest test accuracy. If we compare only the methods involving BP decomposition, SCPL still performs best when using ResNet as the network architecture. We also experiment with BP, Early Exit, AL, and SCPL on CIFAR-10 and CIFAR-100. The results, as shown in Table~\ref{tab:cifar10-cmp} and Table~\ref{tab:cifar100-cmp}, respectively. The results are similar to those on Tiny-ImageNet: SCPL performs best among local objective-based training strategies in all network architectures. However, SCPL performs worse than BP when ResNet is used. These results are consistent with those reported in~\citep{kao2021associated, wu2022associated}.


\subsection{Accuracies of SCPL and BP under different batch sizes} \label{sec:train-batch-size-vs-test-acc}

\figccwidth{lstm-agnews-batchsize-vs-acc.pdf}{A comparison of SCPL's and BP's test accuracies when using LSTM on AG's news with different batch sizes for training. The shaded area represents the mean plus and minus one standard deviation.}{fig:lstm-agnews-bs-vs-acc}{.9\textwidth}

\figccwidth{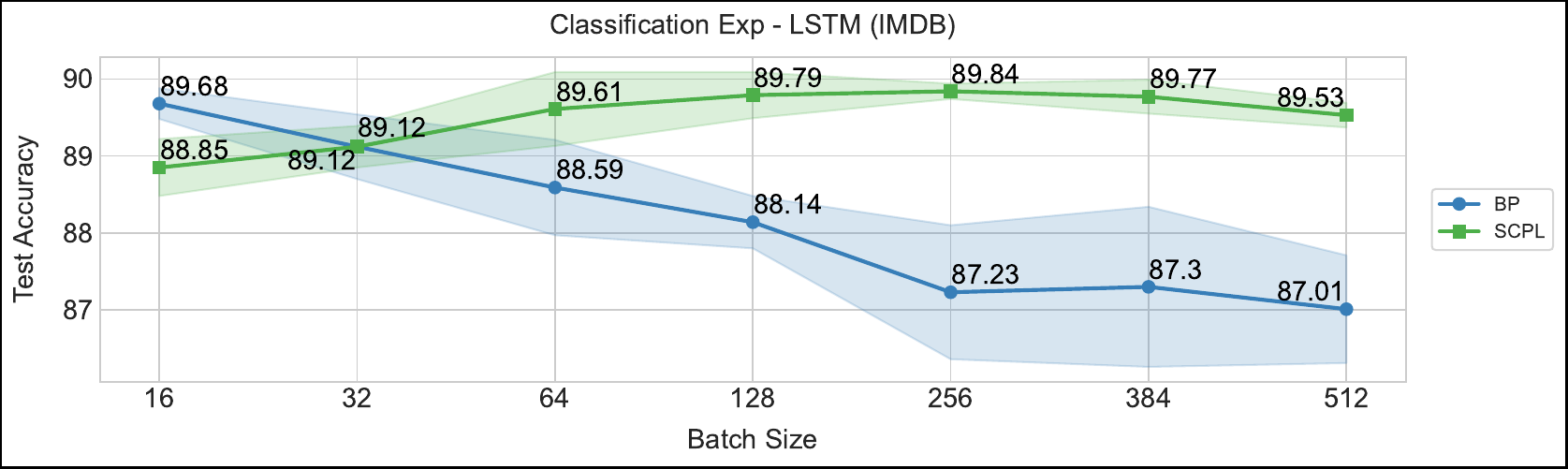}{A comparison of SCPL's and BP's test accuracies when using LSTM on IMDB with different batch sizes for training. The shaded area represents the mean plus and minus one standard deviation.}{fig:lstm-imdb-bs-vs-acc}{.9\textwidth}

\figccwidth{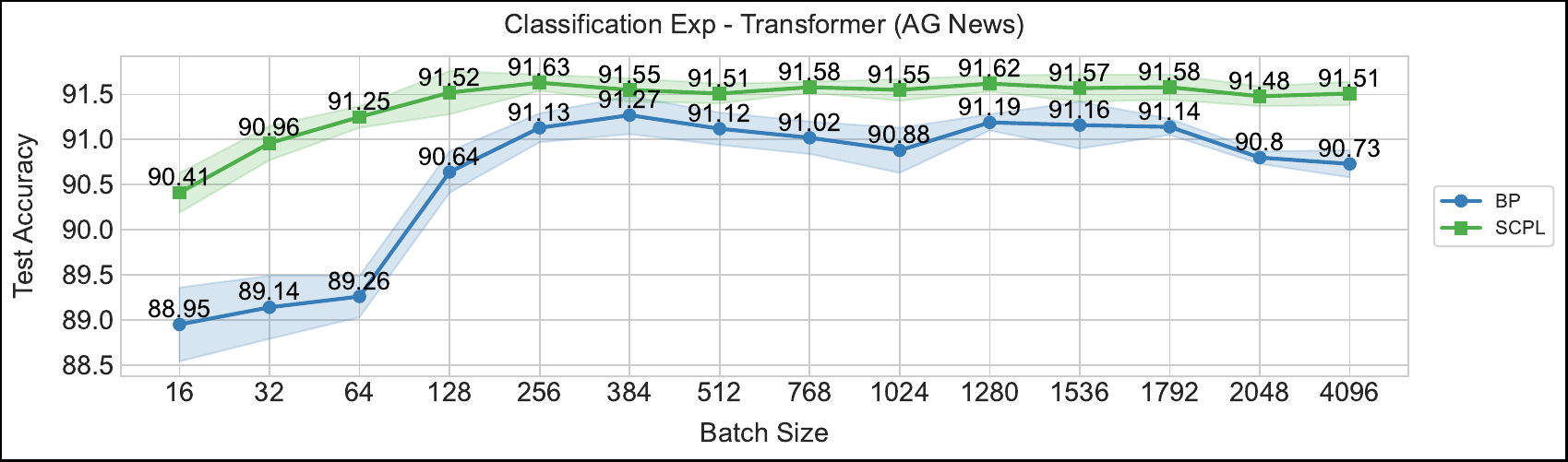}{A comparison of SCPL's and BP's test accuracies when using Transformer on AG's news with different batch sizes for training. The shaded area represents the mean plus and minus one standard deviation.}{fig:transformer-agnews-bs-vs-acc}{.9\textwidth}

\figccwidth{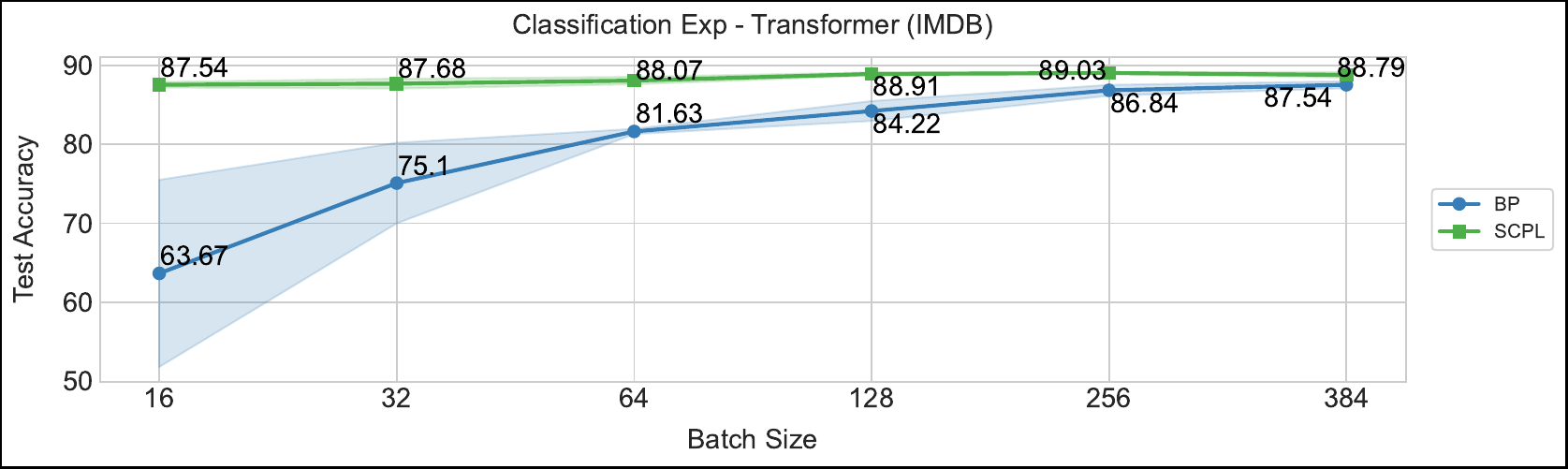}{A comparison of SCPL's and BP's test accuracies when using Transformer on IMDB with different batch sizes for training. The shaded area represents the mean plus and minus one standard deviation.}{fig:transformer-imdb-bs-vs-acc}{.9\textwidth}

This section presents the impact of training batch size on test accuracy. We employed LSTM and Transformer as the backbone networks and conducted experiments using AG's news and IMDB datasets.
The results are shown in Figure~\ref{fig:lstm-agnews-bs-vs-acc}, Figure~\ref{fig:lstm-imdb-bs-vs-acc}, Figure~\ref{fig:transformer-agnews-bs-vs-acc}, and Figure~\ref{fig:transformer-imdb-bs-vs-acc}. Each experiment was carried out five times, and the mean and standard deviation are reported in each figure. The solid lines indicate the mean values, while the shaded areas represent the range within one standard deviation. The results consistently demonstrate that SCPL significantly exceeds BP in test accuracy, evidenced by the minimal overlap between the shaded areas of SCPL and BP.

\subsection{Discussion on accuracy comparison}

When BP is used, all parameters are updated to minimize a global objective -- the residual between the predicted target $\hat{y}$ and the ground truth target $y$. However, methods to decouple end-to-end backpropagation, such as SCPL and AL, are composed of many local objectives, which may differ from the global objective. Therefore, it is surprising that SCPL and AL outperform BP for many network structures. The authors of AL proposed several conjectures to explain this remarkable result. First, projecting the feature vector $\vx$ and the target $y$ into the same latent space, as AL does, may be helpful. Second, the autoencoder components used near the output layers in AL may implicitly perform feature extraction and regularization to some degree. Third, overparameterization may be helpful for optimization~\citep{arora2018optimization, chen2020accelerating}. However, the first and second conjectures apply only to AL but not to SCPL, but SCPL still yields accuracies that are better than those of BP and AL in many cases. Therefore, the above conjectures may not fully explain the success of SCPL. We surmise that, in SCPL, the encoder $f_{\ell}$ in each local component $\ell$ learns to map its input $r_{\ell - 1}^{(i)}$ to a better representation $r_{\ell}^{(i)}$, probably removing noise and outliers and improving model generalization. This is similar to the use cases where contrastive learning is applied to learn better representations of input features in an unsupervised manner. Section~\ref{sec:why-scpl-works} and Section~\ref{sec:scpl-limits} provide a detailed discussion of why it works and the limitations.

The unexpectedly higher accuracy of SCPL compared to that of BP was a pleasant surprise. However, the primary goal of SCPL is to improve training performance. Thus, even in scenarios where SCPL may underperform compared to BP, employing SCPL initially for model training, followed by fine-tuning with standard BP, remains a viable and promising strategy, especially when the size of a model exceeds a GPU's capacity.


\subsection{Training time vs.~test accuracy}


\figwidth{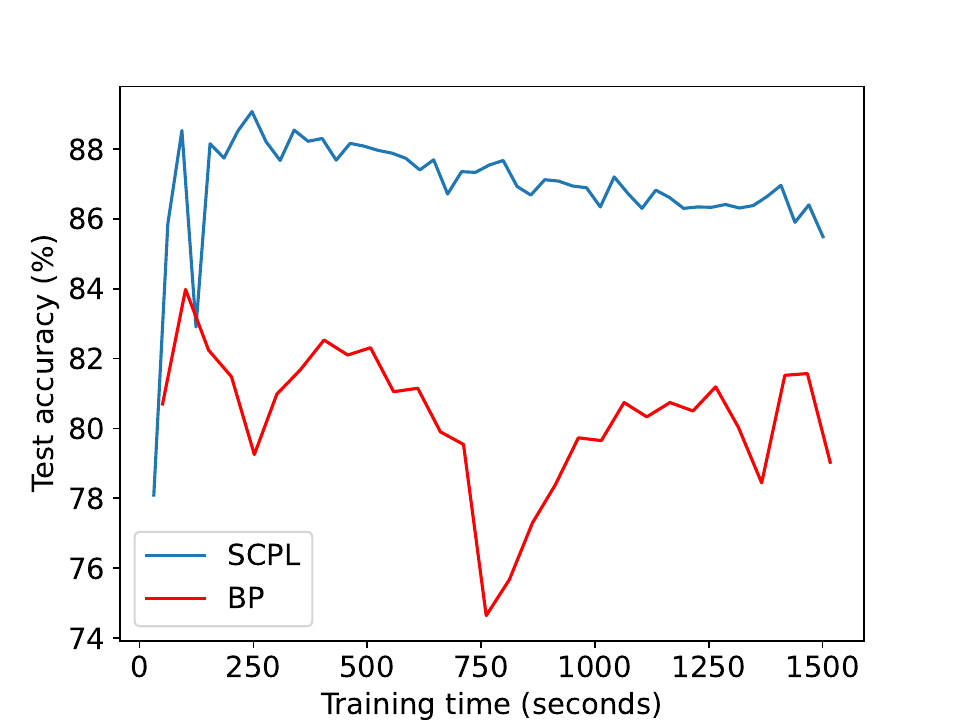}{The relationship between the training time and the test accuracy in the first 30 minutes when using Transformer on IMDB.}{fig:trans-imdb-time-vs-test-acc}{.5\columnwidth}

Figure~\ref{fig:trans-imdb-time-vs-test-acc} shows the training time (the first 30 minutes) and the test accuracy of SCPL (4 GPUs) and BP (1 GPU) in IMDB, using the Transformer as the backbone network. 

Given a fixed training time, SCPL generally achieves better accuracies on test datasets. We do not present the results of BP on 4 GPUs because BP only supports na\"ive model parallelism. Consequently, BP with a single GPU consistently has higher training throughput than BP with 4 GPUs, as each GPU must wait for the others to finish their tasks in both forward and backward, and the additional communication overhead further reduces performance.

Different hyperparameter settings may lead to slightly different curves. However, most of them follow a similar pattern. Experiments on other text datasets (e.g., AG's news and DBPedia) for the Transformer network structure also show similar trends.

\subsection{Profiling NMP and SCPL} \label{sec:profiler}

\figtwo{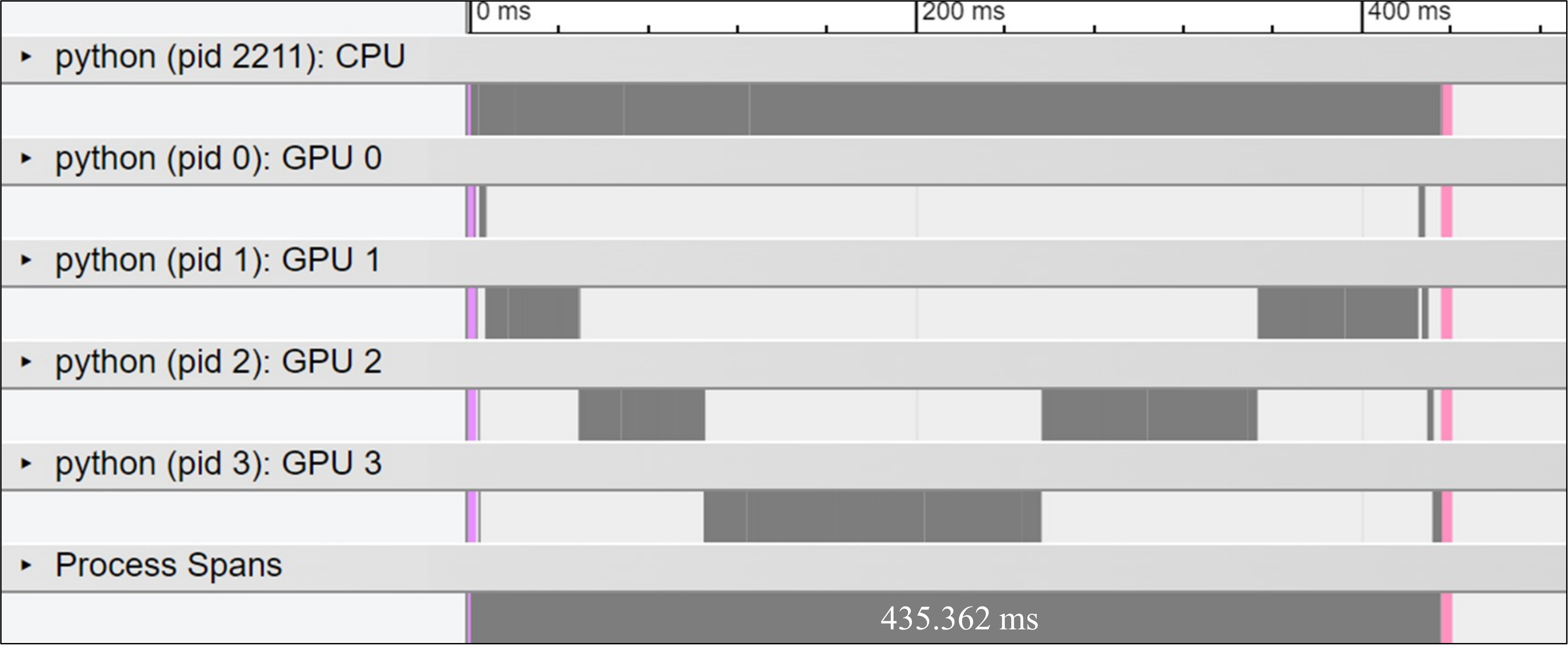}{Training LSTM on IMDB (using NMP).}{fig:lstm-bp-imdb-profiler}
{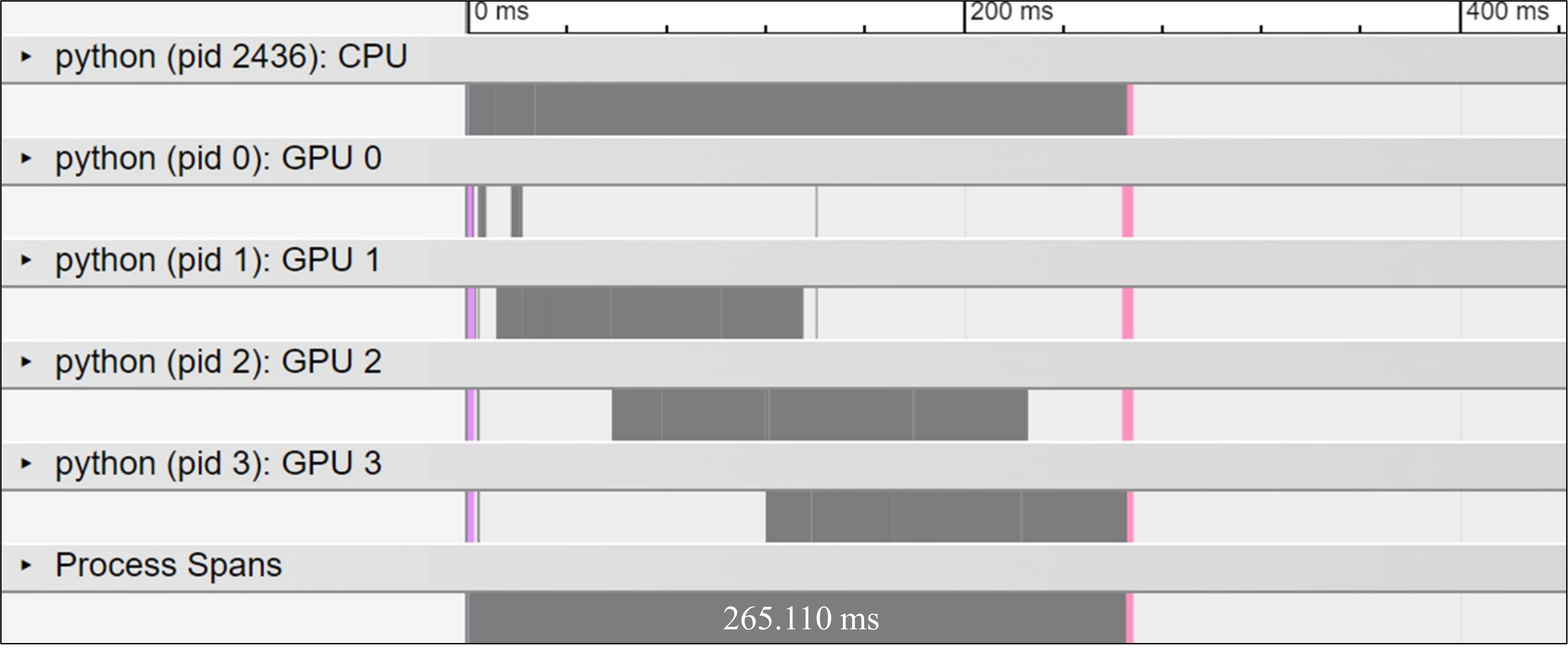}{Training LSTM on IMDB (using SCPL).}{fig:lstm-scpl-imdb-profiler}
{.7\columnwidth}{Visualizing the training job of each device.}
{fig:lstm-imdb-profiler-cmp}

We used PyTorch's profiler to observe the execution periods of the CPU and the GPUs of one iteration. We used 4 GPUs to train an LSTM with four layers; each GPU is responsible for the training of a layer.

Figure~\ref{fig:lstm-imdb-profiler-cmp} shows the CPU's working periods and each GPU's working periods when training by NMP and SCPL. The top row of each subfigure shows the CPU's running periods. Since the CPU handles task scheduling, data preprocessing, data management, non-parallel computation, and job issuing, the CPU runs throughout the training periods in both NMP and SCPL.  

When training by NMP (Figure~\ref{fig:lstm-bp-imdb-profiler}), the GPU0 performs forward for layer 1, then GPU1 performs forward for layer 2, then GPU2 performs forward for layer 3, then GPU3 performs forward for layer 4. GPU3 continues to perform backward for layer 4, GPU2 continues to perform backward for layer 3, GPU1 continues to perform backward for layer 2, and GPU0 continues to perform backward for layer 1. Finally, the CPU asks all GPUs to update the parameters based on the computed gradients (red bars). As shown, all the GPUs perform the forward and backward operations sequentially, causing forward and backward locking, so many bubbles exist among the dependent tasks. The total training time for this iteration is 435.362 ms.

When training by SCPL (Figure~\ref{fig:lstm-scpl-imdb-profiler}), the operations on different GPUs may overlap. In particular, when GPU0 finishes the forward for layer 1, the following operations may co-occur: backward for layer 1 (on GPU0) and forward for layer 2 (on GPU1). Similarly, when GPU1 finishes the forward operation for layer 2, backward for layer 2 (on GPU1) and forward for layer 3 (on GPU2) may co-occur. After GPU2 finishes the forward operation, backward for layer 3 (on GPU2) and forward for layer 4 (on GPU3) may occur concurrently. Finally, GPU3 performs the backward for layer 4, and then the CPU issues an update command for all GPUs (the red bars). Since many bubbles are removed, the total training time for this iteration is reduced to 265.110 ms.

The empirical results match the illustrated example at the bottom of Figure~\ref{fig:bp-vs-scpl}.

\subsection{The effect of batch size and projection head}

\begin{table}[tb]
\caption{The test accuracies of SCPL when using different batch sizes on CIFAR-10.}
\label{tab:batchsize-cmp}
\centering
\begin{tabular}{c||cc}
\toprule
Batch Size & VGG & ResNet\\ 
\midrule
32 & $92.67 \pm 0.10$ & $92.54 \pm 0.14$ \\
128 & $93.11 \pm 0.16$ & $92.53 \pm 0.11$ \\
1024 & $93.42 \pm 0.11$ & $92.78 \pm 0.11$ \\
\bottomrule
\end{tabular}
\end{table}

We also tested how batch size influences test accuracy. As shown in Table~\ref{tab:batchsize-cmp}, performing SCPL training using a large batch size is helpful, and the improvement on VGG is more evident than in other networks. This finding is consistent with the results reported in previous studies, e.g.,~\citep{chen2020simple, henaff2020data, bachman2019learning}, in which the authors noted that because a larger batch tends to include more negative pairs (as shown in Equation~\ref{eq:scl-loss}), the model has access to more information that can be used to distinguish positive pairs from negative pairs. Although other studies, e.g.,~\citep{mitrovic2020less}, have shown that the number of negative pairs may not be critical to improving the test accuracy, most studies tend to agree that a larger batch size leads to better results.

\begin{table}[tbh]
\caption{The test accuracies when using different projection heads on CIFAR-10.}
\label{tab:proj-head-cmp}
\centering
\begin{tabular}{c||cc}
\toprule
Type of projection head & VGG & ResNet\\ 
\midrule
Identity & $79.2$ & $84.2$ \\
Linear & $90.4$ & $90.2$ \\
MLP & $93.0$ & $92.4$ \\
\bottomrule
\end{tabular}
\end{table}

The complexity of a projection head also influences the accuracy, as shown in Table~\ref{tab:proj-head-cmp}, which compares the accuracy of VGG and ResNet in CIFAR-10 using three different projection heads: identity mapping (i.e., $g_{\ell}\left(\vr_{\ell}^{(i)}\right) = \vr_{\ell}^{(i)}$), linear mapping (i.e., $g_{\ell}\left(\vr_{\ell}^{(i)}\right) = \vw^T \vr_{\ell}^{(i)} + w_0$), and the default mapping based on a multilayer perceptron (MLP).  The results are similar to those reported in~\citep{chen2020simple}: the MLP mapping shows an improvement of $2.6\%$ on VGG over the linear projection, which outperforms the identity projection by more than $10\%$ on VGG.

Using an MLP as the projection head is beneficial, probably because the information loss induced by contrastive loss is more severe when a simple projection head is used~\citep{chen2020simple}. In particular, since a projection head $g_{\ell}$ (see Figure~\ref{fig:cl-vs-scl} and Figure~\ref{fig:nn-vs-scpl}) maximizes the agreement between augmented images, $g_{\ell}$ may remove information relevant to image rotation, flipping, and other data augmentation operations, which could be useful for downstream tasks. When using a simple projection head $g_{\ell}$, the information contained in $\vr_{\ell}^{(i)}$ will be similar to that of $\vz_{\ell}^{(i)}$, which means that $\vr_{\ell}^{(i)}$ is invariant to data augmentation. On the other hand, when a complex projection head, such as an MLP, is used, the information in $\vr_{\ell}^{(i)}$ may be very different from that in $\vz_{\ell}^{(i)}$. As a result, even if $\vz_{\ell}^{(i)}$ loses information relevant to data augmentation, $\vr_{\ell}^{(i)}$ may still preserve this information.

\subsection{Ablation Summary of Key Designs and Observations}

\begin{table}[t]
\centering
\caption{\rev{A summary of SCPL components and factors with experiment examples}}
\label{tab:ablation}
\begin{tabularx}{\linewidth}{@{}lXX@{}}
\toprule
Factor / Component & Dataset / Architecture & Key Observation / Metric \\ \midrule
Local Objective & All datasets / all models & SCPL significantly recovers accuracy compared to Early-Exit by preserving features via SCL. \\ \addlinespace
Hardware Scalability & IMDB / Transformer and Tiny-ImageNet / VGG& Training throughput increases sub-linearly with GPUs. \\ \addlinespace
Batch Size & CIFAR-10 / VGG, ResNet & Larger batches provide more negative pairs, essential for refining local representation. \\ \addlinespace
Projection Head & CIFAR-10 / VGG, ResNet & Non-linear MLP heads outperform linear/identity mappings by mitigating information loss. \\ \bottomrule
\end{tabularx}
\end{table}

\rev{This section synthesizes the experimental evidence regarding the key design choices of SCPL. Table~\mbox{\ref{tab:ablation}} provides a comparison of different factors. The primary takeaways from our ablation studies are summarized below.}

\begin{itemize}
    \item \rev{\textbf{Importance of Local Objective}: Replacing standard auxiliary classifiers (Early-Exit) with Supervised Contrastive Loss (SCL) is essential for performance. For example, on IMDB with LSTM, SCPL achieves 89.84\% accuracy compared to 84.34\% for Early-Exit, suggesting that SCL may preserve more task-relevant information.}
    \item \rev{\textbf{Efficiency of Decoupled Pipelining}: SCPL increases hardware utilization by effectively removing the idle bubbles inherent in NMP. In our Transformer/IMDB experiments, scaling from 1 to 4 GPUs results in a throughput speedup from 1.12 times to 1.92 times when the batch size is 32.}
    \item \rev{\textbf{Sensitivity to Batch Size}: The effectiveness of SCPL is highly dependent on the contrastive learning environment. Using a larger batch size provides more negative pairs, which is crucial to refining the discriminative quality of local features.}
    \item \rev{\textbf{Role of Projection Head}: The use of a non-linear MLP projection head acts as a crucial buffer, protecting the main encoder from losing information during local optimization. Switching from identity mapping to an MLP head improved the accuracy of VGG in CIFAR-10 from 79.2\% to 93.0\%.}
\end{itemize}

\subsection{Practitioner Note}

\rev{Based on our empirical evaluations, we offer the following practical recommendations for ones seeking to implement SCPL for their applications.}
\begin{itemize}
    \item \rev{\textbf{Optimal Hyperparameters}: To achieve the best balance between speed and accuracy, we recommend using an MLP projection head rather than a linear or identity mapping, and a larger batch size (e.g., 1024) to ensure a rich set of negative pairs for the local contrastive objective.}
    \item \rev{\textbf{Handling Residual Architectures}: Although SCPL enhances throughput across most tested models, it may exhibit a slight accuracy trade-off in certain deep architectures like ResNet.}
    \item \rev{\textbf{Hybrid Training Strategy}: For applications where peak accuracy is non-negotiable, we recommend a hybrid training approach: utilize SCPL for the majority of the training process to rapidly reach a near-optimal state, followed by a short, end-to-end backpropagation fine-tuning phase to close any remaining accuracy gap.}
\end{itemize}
\section{Discussion} \label{sec:disc}

\subsection{Why SCPL may work?} \label{sec:why-scpl-works}

The experimental results presented in this paper demonstrate that SCPL not only significantly enhances training throughput but also often achieves test accuracies comparable or even superior to traditional end-to-end backpropagation across various architectures and datasets. This phenomenon, where a decoupled training approach with local objectives rivals or exceeds a globally optimized one, warrants a deeper exploration into the potential theoretical underpinnings of SCPL's effectiveness. Although the primary motivation for SCPL design is throughput improvement, its strong empirical accuracy invites further discussion. The authors of Associated Learning (AL) proposed conjectures for similar observations with their method~\citep{wu2022associated}, including the benefits of projecting features and targets into the same latent space, the regularization effect of autoencoder components, and the advantages of overparameterization. While some of these are specific to AL's architecture, the success of SCPL suggests broader principles at play, as the first and second conjectures of AL do not apply to SCPL. We propose several hypotheses to explain the unexpectedly high predictive accuracy of the SCPL method.

First, hierarchical feature refinement via local supervised contrastive learning may be helpful in learning representations. Deep neural networks inherently learn hierarchical features. SCPL's strategy of placing local Supervised Contrastive Learning objectives at intermediate segments can be seen as a way to guide and enhance this process. Each local SCL loss encourages its corresponding network segment to learn discriminative representations by pulling semantically similar samples (based on labels) together in its embedding space while pushing dissimilar ones apart. This ``divide-and-conquer'' approach ensures that each segment actively refines the characteristics for the given supervised signal. As noted above, the encoder $f_l$ in each local component $l$ is driven to map its input $r_{l-1}^{(i)}$ to a more robust and cleaner representation $r_l^{(i)}$, potentially filtering out noise and focusing on task-relevant information at each stage. This iterative refinement process, guided by SCL at multiple points, could lead to a final representation that is perhaps well-structured for the downstream global task.

Second, our design likely improves gradient dynamics and introduces implicit regularization~\citep{chen2019differentiating}. In very deep networks, end-to-end BP can suffer from the issue of vanishing gradients, making it difficult for earlier layers to receive strong learning signals. Local losses, as employed in SCPL, create shorter, more direct paths for gradient flow to pass information to the intermediate segments. This may help learning in earlier layers to develop meaningful representations. Furthermore, the introduction of multiple local objectives can act as a form of implicit regularization. By requiring different parts of the network to satisfy these auxiliary SCL tasks, SCPL imposes additional constraints on the learned representations. This may prevent the model from overfitting to the global objective alone and encourage the development of more generalizable features. This is somewhat analogous to how multitask learning can lead to more robust representations.

Third, although we introduce local objectives, they may mostly align with the global discriminative task. This is mainly because SCL is inherently a supervised task that aims to maximize inter-class variance and minimize intra-class variance in the learned embedding space. Its objectives are conceptually aligned with many global classification tasks, such as those using a final cross-entropy loss.

Finally, the idea that guiding intermediate layers can be beneficial is not entirely new and resonates with earlier paradigms such as layerwise pretraining or the use of auxiliary classifiers in networks (e.g., GoogLeNet~\citep{szegedy2015going}). While SCPL's mechanism is different, it shares the heuristic intuition that ensuring well-formed representations at various depths can lead to better overall models. The overparameterization introduced by the auxiliary projection heads for each SCL loss might also contribute to easier optimization, a phenomenon observed in other deep learning contexts.

In conclusion, while the unexpected high accuracy of SCPL is a welcome outcome, its primary design goal remains the enhancement of training throughput. The theoretical considerations discussed above provide plausible explanations for its strong empirical performance. They suggest that the structured guidance provided by local SCL objectives, coupled with improved gradient dynamics and implicit regularization, can lead to the development of high-quality hierarchical features that benefit the global task. However, understanding the precise theoretical guarantees and limitations of such decoupled learning schemes remains a research issue. 

Even in scenarios where SCPL might slightly underperform BP in final accuracy for specific settings, its utility in accelerating the initial phases of training for very large models, potentially followed by a shorter phase of end-to-end fine-tuning, presents a viable and promising strategy.

\subsection{Potential Risks and Trade-offs} \label{sec:scpl-limits}

Although SCPL demonstrates strong empirical performance, its design implicitly addresses several challenges.

A primary risk is the information loss: the local objective might be too aggressive, forcing a network segment to discard information that, while not critical for the local task, could be valuable for downstream segments or the final global objective. SCPL mitigates this in two ways. First, the use of a non-linear MLP as a projection head is crucial. It allows the intermediate representation $r_l^{(i)}$ to preserve important features even if the projected representation $z_l^{(i)}$ discards them to satisfy the contrastive loss, preventing information loss in the main pathway. Second, the final global loss ensures that the entire network is still incentivized to preserve the features necessary for the end-to-end task.

The second issue is the risk of suboptimal convergence: the greedy nature of optimizing local segments might guide the network toward a solution that is a composite of local optima but not necessarily the global optimum for the end-to-end task. SCPL's design, which integrates a task-specific global loss on the final layer, addresses this trade-off directly. This final integration phase ensures that locally learned, class-discriminative features are ultimately fine-tuned and combined to serve the overall objective, reducing the risk of converging to a suboptimal global solution. In addition, as discussed in Section~\ref{sec:why-scpl-works}, local objectives can implicitly align with the global objective since SCL is inherently a supervised task that aims to maximize
inter-class variance and minimize intra-class variance in the learned embedding space.

Third, SCPL may not be universally optimal for all architectures. Our experiments show that, while SCPL often matches or exceeds BP, it can underperform on ResNet architectures. This is likely because ResNet's inherent skip connections already provide a robust path for gradient flow, diminishing one of the key benefits of local losses. The aggressive local SCL objectives might also interfere with the delicate optimization of residual features that end-to-end BP provides for ResNet. This probably suggests that SCPL is most beneficial in very deep networks that are susceptible to gradient flow issues and may be less advantageous in architectures that have already considered this issue internally.

Finally, SCPL introduces new hyperparameters, including network segmentation points, the temperature of the SCL losses, and the relative weighting between local losses and the final global loss. The optimal tuning of these can be more complex than for standard BP and can influence the final model performance. Moreover, conventional wisdom for tuning hyperparameters (e.g., learning rate, dropout rate) in end-to-end backpropagation models may not be effective in finding the optimal settings for SCPL.

\subsection{SCPL vs.~GPipe: Comparative Analysis of Deployment Efficiency and Throughput} \label{sec:scpl-vs-gpipe}

\figccwidth{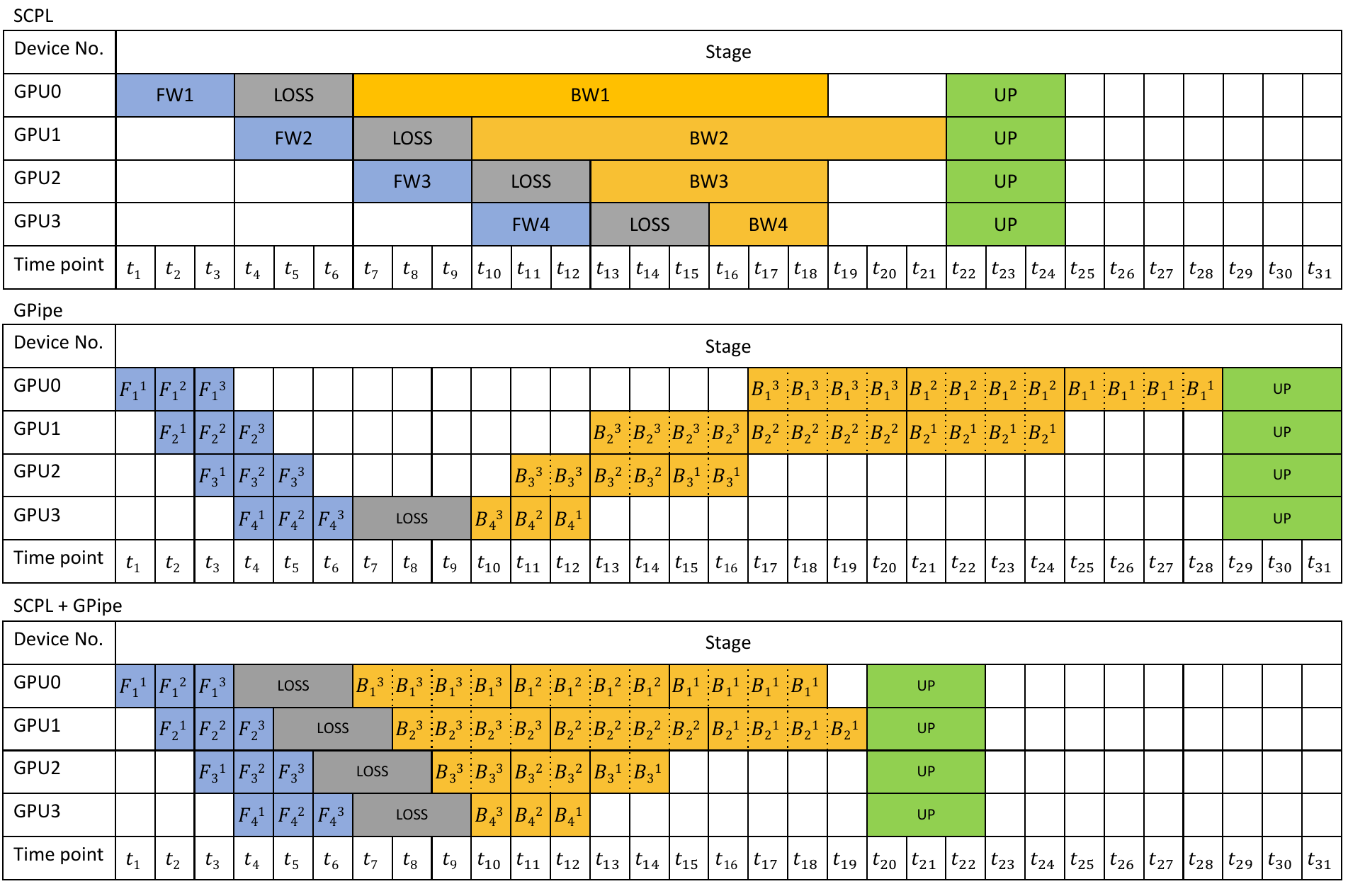}{A comparison of SCPL, GPipe, and integration of both.}{fig:scpl-vs-gpipe}{\textwidth}

\begin{table}[tb]
\centering
\caption{\rev{Theoretical comparison of training iteration costs for the example in \mbox{Figure~\ref{fig:scpl-vs-gpipe}}}}
\label{tab:time_units}
\begin{tabular}{lccl}
\toprule
Method & Time Units & Savings vs. BP & Bottleneck Addressed \\ \midrule
Standard BP / NMP & 51 & 1.00$\times$ & -- (Baseline) \\
GPipe             & 31 & 1.65$\times$ & Forward Locking \\
SCPL (Ours)       & 24 & 2.13$\times$ & Backward Locking \\
SCPL + GPipe      & \textbf{22} & \textbf{2.32$\times$} & Integrated Parallelism \\ \bottomrule
\end{tabular}
\end{table}

\rev{To provide a clear, deployment-oriented benchmark, we quantify the throughput efficiency of different parallelization strategies based on the standardized time units (TU) derived from the non-symmetric workload analysis in Figure~\mbox{\ref{fig:scpl-vs-gpipe}}. In this illustrated scenario, standard end-to-end backpropagation suffers from severe backward locking, resulting in a total sequential cost of 51 TU ($12\text{ FW} + 3\text{ LOSS} + 33\text{ BW} + 3\text{ UP}$, where BW is the sum of $12+12+6+3$ TU across four layers).

Although SCPL and GPipe share architectural similarities in leveraging pipelining, they address different bottlenecks. GPipe tackles \textit{forward locking} by subdividing mini-batches into micro-batches (e.g., $F_{\ell}^1, F_{\ell}^2, F_{\ell}^3$ in the middle subfigure of Figure~\mbox{\ref{fig:scpl-vs-gpipe}}). This overlap allows GPU$i+1$ to begin $F_{\ell+1}^1$ as soon as GPU$i$ finishes $F_{\ell}^1$, reducing the cost to 31 TU. However, GPipe still leaves substantial idle ``bubbles'' during the backward phase.

In contrast, SCPL focuses on \textit{backward locking} by decoupling the gradient flow. This allows different layers to compute parameter gradients simultaneously (refer to the top subfigure in Figure~\mbox{\ref{fig:scpl-vs-gpipe}}), reclaiming idle periods and further reducing the cost to 24 TU. By integrating both (employing micro-batches alongside local objectives), the hybrid approach (SCPL+GPipe) achieves a peak efficiency of 22 TU.

A comparison of NMP, GPipe, SCPL, and an integration of SCPL and GPipe is given in Table~\mbox{\ref{tab:time_units}}.}

\subsection{Managerial and Organizational Implications}

Beyond algorithmic improvements and performance benchmarks, the principles and empirical results of SCPL have significant implications for the management and deployment of AI within organizations from several perspectives.

\rev{First, SCPL improves organizational agility by addressing substantial cost and time barriers to AI adoption. By eliminating the pipeline bubbles inherent in backward locking, SCPL ensures that expensive GPU clusters operate more continuously, fundamentally improving the economic efficiency of hardware resources. As compute costs are projected to increase by 89\%~\mbox{\citep{ibm2024}}, the ability to significantly shorten training durations enables organizations to mitigate these escalating expenses and accelerate time-to-deployment. For the MIS field, this efficiency provides a critical pathway for training and applying sophisticated information systems with greater speed and responsiveness to market changes.}

Second, SCPL functions as an enabling technology for more powerful business intelligence. The performance of many core MIS applications, such as financial fraud detection, supply chain optimization, or advanced recommendation systems, is often correlated with the scale and complexity of the underlying models. By making it more feasible to train giant models that might otherwise exceed the memory of a single device, SCPL provides the technical foundation necessary for developing a new class of more accurate and capable enterprise systems.

Finally, this study offers valuable system-level insights for enterprise IT strategy. Our profiler analysis (Section~\ref{sec:profiler}) comparing GPU utilization between SCPL and Na\"{i}ve Model Parallelism (NMP)  provides a clear, visual argument for why decoupled-loss architectures can be more efficient. This, combined with the architectural discussion of GPipe (Section~\ref{sec:scpl-vs-gpipe}), equips technology leaders with a deeper understanding of the trade-offs involved in different parallelization strategies. These insights are crucial to making informed decisions when designing and investing in scalable and cost-effective AI infrastructure.

\section{Conclusion, limitation, and future work} \label{sec:conc}

This paper proposes to utilize local losses to achieve model parallelism for neural network training. We present SCPL, an innovative methodology for decoupling the components of the backpropagation process in a neural network and accomplishing model parallelism. SCPL applies to all supervised discriminative models, potentially benefiting the training of various large models. Our experiments on multiple open datasets and popular network architectures demonstrate that SCPL reduces training time while achieving test accuracies comparable to or superior to the traditional backpropagation algorithm. Moreover, SCPL can potentially address issues arising from long-gradient flows in deep neural networks. Compared with AL, a state-of-the-art alternative to backpropagation, SCPL is more flexible because it does not require additional fully connected layers near the output layers. This flexibility makes SCPL a promising substitute for AL and an attractive alternative to backpropagation. We believe that SCPL has the potential to advance the field of deep learning and contribute to the development of model parallelism and more efficient alternatives for end-to-end backpropagation.

SCPL has the following limitations that may deserve further study. First, SCPL involves auxiliary parameters, which can lengthen the training time due to their participation in training and require a larger memory footprint. This issue is particularly prominent in the visual domain of SCPL. One way to address this is by using pooling to reduce the size of each block's feature map. Second, forward locking still exists in SCPL, so exploring the possibility of further subdividing forward tasks into smaller micro-batches, similar to GPipe, could be a worthwhile direction; details are as discussed in Section~\ref{sec:scpl-vs-gpipe}. Third, while data parallelism usually does not need any changes to the network architecture, model parallelism typically requires some alterations to the network structure, and SCPL is no exception. This limitation may prevent practitioners from using SCPL. Careful packaging may reduce the barrier. For example, the \texttt{torchpipe} library\footnote{\url{https://github.com/kak brain/torchgpipe}} simplifies the implementation of GPipe.  So, another future work is to implement a library for SCPL to provide a better user interface for developers. We plan to support widely used components in the library, such as fully connected, convolutional, LSTM, and Transformer layers. Fourth, we are also interested in comparing methods that leverage local training losses regarding their training efficiency and test accuracy. Finally, the current SCPL only applies to supervised discriminative models. We plan to redesign SCPL to be applied to generative models like the Generative pre-trained Transformer (GPT).

\section*{GenAI Usage Disclosure}

During the preparation of this work, the authors used large language models (e.g., ChatGPT and Gemini) to improve language and readability. The authors reviewed and edited the content as needed and take full responsibility for the content of the publication.

\section*{Acknowledgments}
We acknowledge support from National Science and Technology Council of Taiwan under grant number 113-2221-E-008-100-MY3.

\bibliographystyle{unsrt}  
\bibliography{ref}

\end{document}